%% file: arxiv.tex
\documentclass{article}

\usepackage{PRIMEarxiv}

\usepackage[utf8]{inputenc}
\usepackage[T1]{fontenc}
\usepackage{cite}
\usepackage{amsmath,amssymb,amsfonts}
\usepackage{graphicx}
\usepackage{booktabs}
\usepackage{multirow}
\usepackage{xspace}
\usepackage{url}
\usepackage{xcolor}
\definecolor{linkblue}{HTML}{1D4ED8}
\definecolor{citegreen}{HTML}{047857}
\definecolor{urlplum}{HTML}{9D174D}
\usepackage{hyperref}
\hypersetup{
  colorlinks=false,
  pdfborder={0 0 1},
  linkbordercolor=linkblue,
  citebordercolor=citegreen,
  urlbordercolor=urlplum,
  filebordercolor=urlplum,
  pdftitle={Pre-Decoding Acoustic Triage for Budgeted Vision-Language Captioning of Untrimmed Egocentric Video},
  pdfauthor={Masoud Jalayer, Changyi Li, Yu Xiao}}
\usepackage{microtype}

\input{tables/macros.tex}

\title{Pre-Decoding Acoustic Triage for Budgeted Vision--Language Captioning
of Untrimmed Egocentric Video}

\author{
  Masoud Jalayer \\
  Dept.\ of Information and Communications Engineering \\
  Dept.\ of Electrical Engineering and Automation \\
  Aalto University, 02150 Espoo, Finland \\
  \texttt{masoud.jalayer@aalto.fi} \\
  \And
  Changyi Li \\
  Dept.\ of Information and Communications Engineering \\
  Aalto University, 02150 Espoo, Finland \\
  \texttt{changyi.li@aalto.fi} \\
  \And
  Yu Xiao \\
  Dept.\ of Information and Communications Engineering \\
  Aalto University, 02150 Espoo, Finland \\
  \texttt{yu.xiao@aalto.fi} \\
}

\begin{document}

\title{Pre-Decoding Acoustic Triage for Budgeted Vision--Language Captioning
of Untrimmed Egocentric Video}

\author{%
\textbf{Masoud Jalayer}\textsuperscript{1,2} \quad
\textbf{Changyi Li}\textsuperscript{1} \quad
\textbf{Yu Xiao}\textsuperscript{1}\\[0.45em]
\textsuperscript{1}Dept.\ of Information and Communications Engineering,
\textsuperscript{2}Dept.\ of Electrical Engineering and Automation\\
Aalto University, 02150 Espoo, Finland\\
\texttt{\{masoud.jalayer, changyi.li, yu.xiao\}@aalto.fi}}

\date{An abridged version of this manuscript is under review at the IEEE
International Conference on Big Data 2026, Industrial and Government Track.
This version restores material cut for the page limit and adds an appendix with
the full sweeps, the per-recording results and the operating points.\\[0.35em]
Code and every results file this manuscript reads:\\
\url{https://github.com/masjalayer/PreDecoding-AcousticTriage} (on request,
pending public release).}

\maketitle

\begingroup\small
\noindent An abridged version of this manuscript is under review at the IEEE
International Conference on Big Data 2026, Industrial and Government Track.
This version restores material cut for the page limit and adds an appendix with
the full sweeps, the per-recording results and the operating points.
\par\endgroup
\vskip 1em

\begin{abstract}
Automatically analyzing hours-long egocentric video is increasingly essential
for progress monitoring, quality control, and safety in logistics,
construction, and manufacturing. Yet current pipelines that process short,
fixed-size windows with a vision--language model (VLM) are prohibitively
expensive because cost scales with the number of model calls. To reduce this
cost, prior work proposes triage policies to select which windows merit a VLM
invocation. However, these policies either sample uniformly or rank windows
using visual features, which ironically requires the video decoding that
the budget constraints are meant to avoid. We propose audio-first triage: select windows using the lightest modality, scored before any video frame is decoded, so the approach composes naturally with token compression or quantization. The novelty lies in the objective, not the representation: rather than a per-frame sound-event detector, we train the selector to trigger once per action. This objective shift improves action coverage by
$\objGainMin$--$\objGainMax$ percentage points across all evaluated call
rates, using frozen AudioSet-pretrained features without domain-specific
sound-event labels. Using fewer than half of the available calls, the triage
cuts $\saveMin$--$\saveMax\%$ of VLM calls at matched coverage on
EPIC-KITCHENS-100 (EK-100), surpasses uniform sampling through the mid-range on Ego4D over
\egoSelClips\ clips, and outperforms two recent visual keyframe selectors.
Code, the reference implementation and every results file this manuscript reads
are at \url{https://github.com/masjalayer/PreDecoding-AcousticTriage}.
\end{abstract}

\section{Introduction}

Automatically analyzing hours-long egocentric video is increasingly useful for progress monitoring, quality control and safety in logistics, construction and manufacturing. Wearable cameras suit these settings: they capture manipulation from the actor's viewpoint, and a single device can record an entire shift without instrumenting the environment. The challenge is economic, not sensory: recordings are continuous, but the compute budget is not. A VLM does not ingest a recording end-to-end; it is called repeatedly on short windows, and cost scales with the number of calls. Forwarding every four-second window of a 1{,}000-hour archive to an 8B model costs on the order of \gpuFull\ GPU-hours before any downstream analysis. This creates a critical gap: we need principled triage policies that decide which windows merit a call, maximizing coverage of meaningful events while minimizing compute. The operative question is therefore \textit{which windows to spend calls on, and how to select them efficiently under strict compute budgets}. 

Unlike content-aware selectors that rank windows by visual features~\cite{tang2025aks, liu2025bolt}, which require the very video decoding the budget aims to avoid, we instead triage using audio, a lightweight modality that can be scored before any frame is decoded. Audio suits this decision: many manipulations are acoustically distinctive, and audio is orders of magnitude cheaper to process than video and naturally compatible with downstream token compression or quantization.

\begin{figure}[t]
\centering
\includegraphics[width=\linewidth]{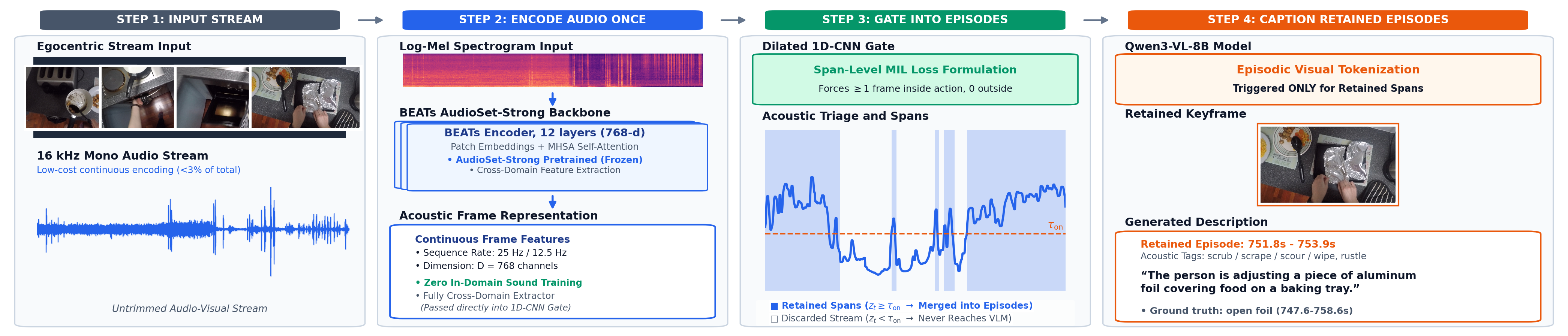}
\caption{The evaluated pipeline: a frozen extractor encodes the audio once, a span-level gate emits episodes, and only those episodes reach the VLM.}
\label{fig:pipeline}
\end{figure}

As illustrated in Fig.~\ref{fig:pipeline}, we introduce a pre-decoding, audio-first triage pipeline that decides which windows of an egocentric stream merit a VLM call before any pixels are decoded. Frozen AudioSet-pretrained features~\cite{chen2022beats} feed a small head trained to trigger once per action, and the highest-scoring windows are then selected under a spacing constraint that keeps successive calls apart. Because both stages run entirely on audio, the method is orthogonal to, and composes with, standard VLM efficiency techniques such as visual token compression or weight quantization. 
The scientific novelty of our proposal lies in aligning the selector’s training objective with the deployment criterion. The natural baseline, a frame-level sound event detector optimized with cross-entropy, overemphasizes marking every frame of long actions, increasing cost without improving recall. We instead use span-level multiple-instance learning (MIL), requiring at least one positive per action and silence elsewhere. With the same AudioSet features, this objective shift more than doubles the amount of stream discarded at fixed recall, overturning the impression (from thresholded posteriors) that acoustic screening is ineffective. Combined with budget-aware spacing, the result is a simple, composable triage mechanism that improves action coverage per call without touching the VLM.

\textbf{Key contributions of this paper are summarized below.}
\begin{itemize}
  \item \textbf{A gate trained for budget rather than for detection.} 
  We train the acoustic frontline with a span-level MIL objective over strong action spans, rewarding one firing inside an action and penalizing the rest. This inverts the usual use of MIL in audio, where weak clip labels supervise dense frame predictions; here strong spans supervise a deliberately sparse predictor, because sparsity is what a call budget pays for. With features, architecture and seed fixed, it improves action coverage by $\objGainMin$ to $\objGainMax$ points over frame-level cross-entropy at every budget (Section~\ref{sec:gate}). 
  \item \textbf{Spacing constraint.}
  We introduce a simple but novel spacing constraint: enforce a minimum separation between calls, chosen on held-out recordings, to counter temporal correlation and convert scores into distinct-action coverage. This scheduler recovers $\selBest\pm\selBestSd\%$ of actions versus $\selBestUnif\%$ at only $\selBestCost\%$ of calls, and outperforms two recent visual selectors using the same scores while operating below half the budget (Section~\ref{sec:select}). The gain over uniform sampling holds on \pairWinsUnif\ of \pooledVids\ held-out EK-100 recordings and persists through the mid-budget range on \egoSelClips\ Ego4D clips (Section~\ref{sec:repl}). 
  \item \textbf{What the evidence supports, and what it does not.}
  Costing triage in calls rather than in retained seconds changes which methods look good, and undercharges our own gate by $\costUndercount\times$ (Section~\ref{sec:problem}). Occupancy bounds what any frontline can add before one is chosen, and every dataset we could measure sits between $\occPubMin$ and $\occPubMax\%$ (Section~\ref{sec:density}), which is why uniform sampling is a strong baseline here and why we mark the budgets at which triage stops paying.
\end{itemize}

The rest of this paper is organized as follows. Section II reviews the related work, followed by the proposed method in Section III. Section IV presents the experimental setup and results, while Section V discusses the main findings and limitations. Finally, Section VI concludes the paper.
\section{Related Work}
\label{sec:related}

\noindent\textbf{Egocentric video understanding.}
Egocentric video understanding spans activity recognition, temporal
localization, and multimodal reasoning. Early work combined first-person
video with inertial sensing under limited supervision~\cite{huang2020knowledge},
while EPIC-KITCHENS-100 (EK-100)~\cite{damen2018epic} and
Ego4D~\cite{grauman2022ego4d} established large-scale benchmarks for
first-person activity understanding. Recent methods increasingly apply large
multimodal models to egocentric video and reduce inference cost through
techniques such as visual-token compression~\cite{luo2025openmmego}.
Temporal localizers, including BSN~\cite{lin2018bsn} and
ActionFormer~\cite{zhang2022actionformer}, operate on decoded or precomputed
visual features, with subsequent work improving representation
capacity~\cite{liu2024adatad} or localizer efficiency~\cite{sun2026liquidtad}.
Our problem precedes these stages: deciding which temporal windows should
undergo visual processing at all.

\noindent\textbf{Sound event detection and MIL.}
AudioSet pre-training has produced strong representations for sound event
detection (SED). PANNs/CNN14~\cite{kong2020panns} established convolutional
AudioSet models as general audio taggers, followed by Transformer-based
architectures such as AST~\cite{gongast}, PaSST~\cite{koutini22passt}, and
HTS-AT~\cite{chen2022htsat}, as well as BEATs~\cite{chen2022beats}.
AudioSet-Strong further improves temporally localized
representations~\cite{schmid2024multi}, while recent work explores
prototype-based masked modeling~\cite{10889422} and open-vocabulary sound
recognition~\cite{cai2025detectsoundopenvocabularysound}. These models are
primarily optimized for frame- or clip-level detection. We instead use audio
as a pre-decoding gate and study supervision aligned with distinct-action
coverage under a limited number of VLM calls. Our experiments compare against
PANNs and several AudioSet-Strong representations, including BEATs,
ATST-Frame~\cite{schmid2024multi}, and fPaSST~\cite{koutini22passt}.

MIL is widely used in weakly supervised audio tagging, where clip-level labels
supervise temporally distributed predictions and the pooling rule controls
evidence aggregation. Wang et al.~\cite{wang2019pooling} compare several
pooling strategies, showing that max pooling concentrates supervision on the
strongest instance whereas smoother alternatives distribute gradients more
broadly. Our setting uses strong action spans as positive bags and requires
evidence somewhere within each span rather than positive predictions
throughout it. This matches the downstream criterion, where one selected
window can already cover an action.

\noindent\textbf{Frame selection under call constraints.}
Visual frame selection addresses a related allocation problem. Adaptive
Keyframe Sampling~\cite{tang2025aks} balances relevance and temporal coverage,
while MDP3~\cite{sun2025mdp3} promotes diversity with determinantal point
processes, building on earlier work such as seqDPP~\cite{gong2014seqdpp}.
Training-free methods pursue similar goals under a fixed frame count:
BOLT~\cite{liu2025bolt} uses inverse transform sampling to distribute frames
according to relevance, and Q-Frame~\cite{zhang2025qframe} performs
query-aware selection with adaptive per-frame resolution. Frame-Voyager
~\cite{yu2025framevoyager} instead trains a selector against the downstream
Video-LLM by ranking frame combinations according to their effect on model
loss.

These methods derive their selection signal from visual observations or
features and therefore incur video-side processing before selection is
complete. Audio can move this decision earlier. Chapter-Llama
~\cite{ventura2025chapterllama} uses speech transcripts to select frames for
captioning, whereas our gate relies on non-linguistic audio and does not
require speech transcription. VFSTA~\cite{wang2025audioguided} is more closely
related, using audio-event cues to guide keyframe selection for description
generation. It nevertheless combines audio with visual features in a
cross-modal model and optimizes description similarity, whereas our selector
operates before visual feature extraction and is evaluated under an explicit
VLM-call limit.

\noindent\textbf{Audio--visual localization.}
Audio and vision are also jointly modeled for temporal and spatial
localization. Audio-visual event localization identifies events that are both
audible and visible~\cite{tian2018ave}, weakly supervised audio-visual parsing
separates audible and visible event components~\cite{tian2020avvp}, and
audio-visual segmentation localizes sound-producing image
regions~\cite{zhou2022avs}. These methods optimize localization or
segmentation using decoded visual input. Our setting instead uses audio before
visual decoding to determine where downstream visual reasoning should be
invoked.

\noindent\textbf{Reducing VLM inference cost.}
A complementary line of work reduces the cost of each VLM invocation rather
than the number of invocations. Video-language models such as
Video-LLaVA~\cite{lin2023videollava} and Qwen-VL~\cite{bai2023qwenvl}
typically process sampled visual inputs, while recent efficiency methods
reduce cost by pruning or merging visual
tokens~\cite{luo2025openmmego,shang2025prumerge} or skipping redundant
computation~\cite{wu2024videollmmod}. These approaches are complementary to
pre-decoding triage: they reduce per-call cost, whereas our method reduces the
number of calls. Section~\ref{sec:compose} evaluates the two together.

Overall, prior work either selects after visual processing, combines audio
with decoded visual features, or optimizes audio models for detection rather
than coverage under limited calls. We instead score the stream from audio
before video decoding, train the gate for distinct-action coverage, and
allocate the resulting scores under an explicit VLM-call constraint.

\section{Method}
\label{sec:method}

\subsection{Budgeted action coverage}
\label{sec:problem}

Consider a recording of duration $T$ with $N$ annotated action intervals $\mathcal{A}=\{A_i\}_{i=1}^{N}$, $A_i=[a_i,b_i)$. A VLM is not given the recording; it is called on windows. Writing $w$ for the temporal support of one call, an exhaustive scan costs $M=\lceil T/w \rceil$ calls, the last window padded. A triage policy retains windows $\mathcal{C}=\{C_j\}_{j=1}^{K}$, and we measure normalized cost and action coverage as
\begin{align}
\operatorname{Cost}(\mathcal{C}) &= \frac{|\mathcal{C}|}{M}, \label{eq:cost}\\
\operatorname{Cov}(\mathcal{C}) &= \frac{1}{N}\sum_{i=1}^{N}
  \mathbb{I}\!\left[\exists\, C_j \in \mathcal{C}:\; C_j \cap A_i \neq \emptyset\right].
\label{eq:coverage}
\end{align}
For a budget $\rho \in (0,1]$ we set $K_{\max}=\lfloor \rho M\rfloor$ and evaluate
\begin{equation}
\max_{\mathcal{C}} \operatorname{Cov}(\mathcal{C})
\quad \text{s.t.} \quad |\mathcal{C}| \le K_{\max}.
\label{eq:cov_objective}
\end{equation}
This is both the training and the evaluation objective. The intervals $\mathcal{A}$ score the result and are never available to the selector at inference; all methods are compared at matched $\rho$, and an action no call touches is lost for good.

We count calls because calls are interchangeable: every window has length $w$ and every call the same frame count, token allowance and prompt, so the total is linear in $|\mathcal{C}|$. We measure the constant instead of assuming it: a call costs \gpuPerCall\,GPU-seconds at our operating point, so the \gpuFull\ GPU-hours of an unfiltered scan, and the cost of any triaged run, follow from the call count alone. It is also the unit the neighboring literature uses, since frame-selection methods are compared at a fixed frame count~\cite{tang2025aks, sun2025mdp3}. Where decoding or bandwidth is the constraint, retained duration is the right unit and we report it separately.

The unit matters more than it appears. A duration-based cost assumes retained time is contiguous and billable in proportion, flattering any rule that emits short fragments: our gate emits \nEpisodesDeployed\ episodes averaging \meanEpisodeS\,s and issues calls on \nCallsDeployed\ windows, where a duration account charges the equivalent of \nDurEquiv, an undercount of $\costUndercount\times$. The objective below makes the gate as sparse as it can be, so the units disagree where the method is most aggressive.

The formulation exposes the objective mismatch. Coverage is invariant to \emph{how much} of an action is reached beyond the first call, while cost is linear in calls, so a cross-entropy classifier rewarded for labeling every frame positive pays for all but one of them. The gate should be as sparse as it can while still touching every action, which is a MIL objective. It exposes a second requirement too: coverage counts \emph{distinct} actions, so ranking by a temporally correlated score wastes calls, and selection must spread as well as score (Section~\ref{sec:select}).

\subsection{Audio representation}
\label{sec:audio_repr}

Audio is resampled to 16\,kHz, the rate the encoder was trained at and enough for the impact and friction transients that characterize manipulation, and encoded by a frozen BEATs backbone~\cite{chen2022beats} with its AudioSet-Strong frame-level head. Writing $u_t$ for the timestamp of feature index $t$, the frontend emits $\mathbf{h}_t \in \mathbb{R}^{447}$, the frame-level posteriors over the AudioSet-Strong vocabulary, which the gate reads in logit space. These are posteriors used as features, not decisions: nothing is thresholded here, and Section~\ref{sec:transfer} shows the objective lifts every extractor we tested to indistinguishable gating performance, so the extractor is not what determines whether the gate works.

Long recordings are processed in $10$\,s chunks with a $5$\,s hop, which gives enough context to disambiguate sustained events while keeping activations localized; overlapping contributions are averaged onto one global frame grid, so each timestamp appears exactly once and the gate consumes frames at \gateRate\,Hz.

The encoder stays frozen: the supervision available is a few dozen annotated recordings, far too little to adapt a full transformer, and freezing keeps the frontline cheap enough for the cost condition of Section~\ref{sec:practice}. In-domain supervision is limited to action spans, the annotation a practitioner already owns as soon as they can say what counts as an action. A frontline needing a domain-specific sound taxonomy first is of little use at a new site, so we treat cross-domain operation as a requirement, not a convenience.

The domain argues for the same objective. Annotating the EK-100 audio independently of the video, Huh et al.~\cite{huh2023epicsounds} find acoustic and visual onsets routinely disagree: closing a bin is visually underway when the hand reaches the lid but silent until it strikes. Reproducing the visual envelope is misspecified twice, once because a single instant suffices and once because the evidence does not occupy that envelope. A span-level criterion is indifferent to both.

The AudioSet head is retained only to produce readable tags for the VLM prompt. An optional segmentation head, a BEATs--SlowFast model~\cite{kazakos2021slowfast} fine-tuned on the deployment domain's 42-class taxonomy, refines boundaries where needed (Section~\ref{sec:seg}) but is no part of triage: gating tolerates saturating coverage, since only the union of spans matters, whereas boundaries are read off transitions, and at a threshold of $0.95$ the AudioSet-Strong extractors place at most $422$--$444$ against $2{,}015$ from a domain-tuned head.

\subsection{Coverage-aligned acoustic gate}
\label{sec:method_gate}

A three-block dilated temporal CNN with dilations $1,4,16$ and $0.41$\,M parameters maps $\{\mathbf{h}_t\}$ to frame-level activity logits $z_t$. The head is deliberately small: gating is a low-dimensional decision over an already-informative representation, and a larger one would raise frontline cost without touching the objective mismatch Section~\ref{sec:gate} identifies as binding.

Write $\mathcal{P}_i=\{t: u_t \in [a_i,b_i)\}$ for the frames inside action $i$ and $\mathcal{N}$ for those inside no action. Frame-level supervision treats every $t \in \mathcal{P}_i$ as independently positive, although (\ref{eq:coverage}) asks only that one window cover the action. We therefore pool the logits within each span with an \emph{unnormalized} log-sum-exp,
\begin{equation}
g_{\tau}(\mathcal{P}_i)=\frac{1}{\tau}\log \sum_{t \in \mathcal{P}_i} e^{\tau z_t},
\label{eq:lse}
\end{equation}
a smooth maximum over the span, with $\tau=4$. Gradient reaches frame $t$ in proportion to $e^{\tau z_t}$, so a large $\tau$ sends it to one frame per span and trains unstably, while a small $\tau$ spreads it evenly and restores the frame-level behavior we are trying to avoid. We do not divide by $|\mathcal{P}_i|$: a mean over the span is exactly the frame-level target (\ref{eq:coverage}) does not ask for.

With $\ell_{\mathrm{BCE}}$ the binary cross-entropy with logits, the gate is trained on
\begin{equation}
\mathcal{L}_{\mathrm{gate}}=
\underbrace{\frac{1}{N}\sum_{i=1}^{N} \ell_{\mathrm{BCE}}\big(g_{\tau}(\mathcal{P}_i),1\big)}_{\text{one frame fires per action}}
+\underbrace{\frac{1}{|\mathcal{N}|}\sum_{t \in \mathcal{N}} \ell_{\mathrm{BCE}}(z_t,0)}_{\text{fire nowhere else}},
\label{eq:gate_loss}
\end{equation}
the two terms weighted equally. The positive term saturates once any frame in the span is confident, so the model gains nothing from widening a detection; the negative term charges it for every frame elsewhere. This is a direct surrogate for maximizing (\ref{eq:coverage}) at fixed (\ref{eq:cost}).

At inference $p_t=\sigma(z_t)$ is median-filtered, removing isolated frame flips without displacing the onsets a mean filter would smear, and turned into intervals by hysteresis: open at $p_t \ge \theta_{\mathrm{on}}$, closing at $p_t < \theta_{\mathrm{off}} \le \theta_{\mathrm{on}}$. This prevents the fragmentation a single threshold produces when the score oscillates about it, each fragment carrying the fixed overhead of a call. Spans below $0.15$\,s are dropped, being shorter than any annotated action, and gaps below $0.6$\,s are closed, a sub-second silence being more cheaply retained than resolved. Both thresholds and the filter width are selected on development recordings (Section~\ref{sec:setup}). The surviving intervals are the \emph{episodes} the end-to-end pipeline captions.

\subsection{Spending the calls}
\label{sec:select_method}

Episodes are what the pipeline describes, but not the unit calls are spent in, and a rule ranking episodes could not be compared with uniform sampling on equal terms. Selection runs on the same $M$-window grid the cost is counted on. Window $W_m$ takes the largest gate score it contains, $s_m=\max_{t:\,u_t \in W_m} p_t$, and we traverse windows in descending $s_m$, retaining $m$ only if
\begin{equation}
|m-k| \ge \delta \quad \forall\, k \in \mathcal{S},
\label{eq:separation}
\end{equation}
where $\mathcal{S}$ holds the windows already taken and $\delta$ is a minimum separation measured in windows. Selection stops at $|\mathcal{S}|=K_{\max}$. We use $\delta=2$, that is $8$\,s, chosen on development recordings. The gap-closing of Section~\ref{sec:method_gate} merges brief silences \emph{within} one episode; $\delta$ separates \emph{between} calls, so the two act at different stages.

The grid is what makes the comparison fair: uniform sampling, AKS and BOLT receive the same $K_{\max}$ windows of the same length and differ from us only in which they take.

\subsection{Visual decoding and VLM inference}
\label{sec:vlm}

Each selected window is one call. Its highest-scoring instant and the top acoustic tags over it go to Qwen3-VL-8B-Instruct~\cite{bai2023qwenvl} with a fixed visual token allowance. Nothing outside the selected windows is decoded, which is where the saving is realized; Section~\ref{sec:negative} reports what happens when a hand-centered crop replaces this input.

\section{Experimental Results}
\label{sec:setup}

\subsection{Data, setup and metrics}
\label{sec:data}
Operational archives of body-worn recordings are rarely public, and those that exist seldom carry both modalities and the frame-accurate action annotations this evaluation requires. We therefore evaluate on two public egocentric datasets bracketing the conditions a deployed system meets: one densely annotated and procedurally structured, the other unconstrained and acoustically diverse. Neither is the object of study, and the cost model of Section~\ref{sec:practice} depends only on quantities an operator can measure on their own archive.

\textbf{EPIC-KITCHENS-100} (EK-100)~\cite{damen2018epic} records unscripted kitchen activity from head-mounted cameras, densely annotated with start and end times for every manipulation. It is the primary benchmark because a frontline can only be scored where the ground truth states exactly when activity occurs. Its accompanying EPIC-SOUNDS annotations~\cite{huh2023epicsounds} supply the acoustic taxonomy used by the optional segmentation head of Section~\ref{sec:seg} and the in-domain ablation; the gate does not use them. We evaluate on \epicEvalVids\ validation recordings spanning \epicParticipants\ participants and \epicAudioHours\ hours, containing \epicActions\ annotated actions, and replicate the selection results on \replVids\ further validation recordings from \replVids\ participants this set never touches (Section~\ref{sec:repl}). The gate is fitted on action annotations from \gateTrainVids\ further training-split recordings, so no recording used for fitting appears in evaluation.

The experiments differ in scope because they differ in what they cost to run. Audio-only results (gating, the objective and extractor comparisons, boundary alignment, throughput) use all \epicEvalVids\ recordings; results needing a VLM call per stimulus (captioning, the token-allowance sweep, the composition analysis, and the deployed-frame check of Section~\ref{sec:placement}) draw \capStimuli\ stimuli from three of them. Conclusions about gating therefore rest on the full evaluation set, those about naming quality on the narrower one.

\begin{figure}[t]
\centering
\centering\includegraphics[width=0.78\linewidth]{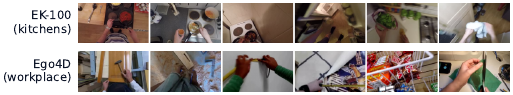}
\caption{Sample frames from the two evaluation datasets: EK-100 kitchens, and Ego4D clips annotated as manual or workplace activity.}
\label{fig:datasets}
\end{figure}

\textbf{Ego4D}~\cite{grauman2022ego4d} covers unconstrained daily and occupational activity, and tests whether a gate tuned on one acoustic domain transfers to another without adaptation, the situation faced at a new site. We use the EgoSound subset~\cite{zhu2026egosound} (\egoClips\ clips, \egoHours\,h), whose annotations are audio question-answering items carrying the timestamp each question refers to, not action intervals; we therefore report retention of \emph{question-referenced acoustic moments} and do not present it as action recall. Of 2{,}346 annotations, 7 have unparseable timestamps and are excluded.

\noindent\textbf{Frameworks compared.}
Table~\ref{tab:methods} lists every system we run end to end. AKS and BOLT are
re-run on our acoustic scores, since the selection stage of each consumes an
arbitrary score sequence. MDP3~\cite{sun2025mdp3}, Q-Frame~\cite{zhang2025qframe} and
Frame-Voyager~\cite{yu2025framevoyager} are not: they need
visual features, a text query, or the downstream model's own loss, none of
which exist before a frame is decoded.

In detail, \emph{full scan} forwards every segment. \emph{Uniform subsampling}
keeps one window in $k$. \emph{Random spans} places spans of matched length
statistics at random, which controls for any generic temporal prior rather than
for the score itself. \emph{PANNs-gated} uses CNN14-DecisionLevelMax with its
published AudioSet checkpoint. \emph{Visual-gated} uses frame differencing and
Farneb\"ack optical flow. Two settings matter for fairness. The pretrained
acoustic detectors are used class-agnostically, with non-manipulation
categories excluded exactly as for our own gate, and max-pooled to our frame
rate so temporal resolution is not a confound. The ego-compensated flow variant
subtracts the global median flow vector before measuring what local motion
remains.

\begin{table}[t]
\centering
\small
\caption{Systems compared, and the signal each one ranks windows by.}
\label{tab:methods}
\input{tables/tab_methods.tex}
\end{table}

\noindent\textbf{Compute environment.}
All experiments run on a shared university HPC cluster under SLURM. Each job is
allocated one GPU, six CPU cores and 64\,GB of host memory; jobs are scheduled
across NVIDIA V100 (32\,GB), A100 (80\,GB) and H100 (80\,GB) nodes. Acoustic
feature extraction, gate training and the VLM stage all run on a single GPU,
and no experiment in this paper requires more than one.

Because the cluster is heterogeneous, timings are reported on a fixed device,
not on whichever node a job happened to receive: throughput figures and
GPU-hour projections are measured on the V100, the slowest of the three, so
they are conservative. An identical configuration rerun on a different GPU can
move a number materially, which is why every comparison is repeated across
seeds and no single run is reported alone. Gate training costs roughly five
minutes per seed, so the whole seed-controlled protocol fits in under half an
hour of GPU time; the captioning runs dominate the cost of the study.

\noindent\textbf{Metrics.}
Cost and coverage are as defined in Section~\ref{sec:problem}; coverage uses non-zero overlap, since a gate's failure mode is discarding an action entirely, not localizing it loosely. Boundary $F_1$ at tolerances $\{0.5,1.0,2.0\}$\,s uses greedy nearest-first matching with each ground-truth boundary consumed once, and is only ever compared at matched segment counts, because $B\text{-}F_1$ rises monotonically with the number of boundaries emitted. Event-level SED uses the full 42-class taxonomy with a $0.5$\,s onset collar, wider than the $200$\,ms DCASE convention~\cite{mesaros2016metrics} because the segmentation head's \frameRate\,Hz frame rate would otherwise make the metric a measure of quantization. Captioning is scored against the dataset's own noun and verb vocabularies; we report no factuality score, as that needs human judgment~\cite{rohrbach2018chair}.

\subsection{Occupancy of the stream}

\begin{figure}[t]
\centering
\includegraphics[width=\linewidth]{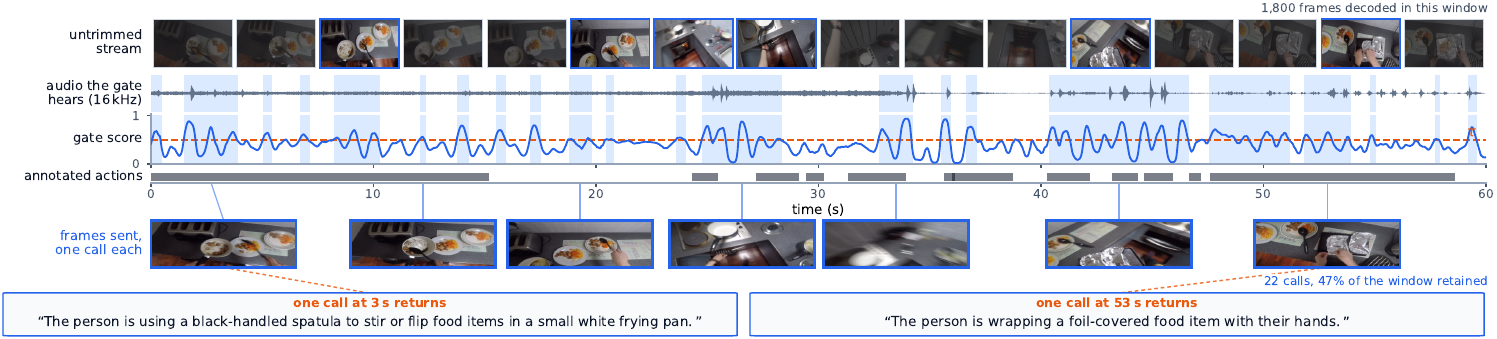}
\caption{An example of sixty seconds of stream at the deployed operating point: decoded frames, the audio, the gate score against $\tau$, the annotated actions, and the frames that survive selection, one call each.}
\label{fig:qualitative}
\end{figure}

Fig.~\ref{fig:qualitative} shows the decision on one stretch of stream. Two properties recur: retained episodes are short and numerous rather than a few long blocks, so the cost unit matters, and the actions beneath them are nearly continuous, leaving uniform sampling hard to beat.

\label{sec:density}

Before comparing filters it is worth asking how much room a filter has. If nearly every window contains something worth describing, then any way of choosing windows, including choosing blindly, reaches most of the actions. We call the fraction of windows holding at least one annotated event the \emph{occupancy} of a dataset,
\begin{equation}
\operatorname{Occ}=\frac{1}{M}\sum_{m=1}^{M}
\mathbb{I}\Big[W_m \cap \big(\textstyle\bigcup_{i} A_i\big) \neq \emptyset\Big],
\label{eq:occupancy}
\end{equation}
computable from annotation files alone. On EK-100 at $w=4$\,s it is $\occupancy\%$, with $\actionsPerWindow$ actions per window on average, so uniform sampling is close to optimal and a frontline has little room to beat it.

\begin{table}[t]
\centering
\small
\caption{Occupancy of each dataset: the fraction of four-second windows holding at least one annotated event.}
\label{tab:occupancy}
\input{tables/tab_occupancy.tex}
\end{table}

This could be a property of kitchens, so we repeated the measurement wherever annotations make it computable (Table~\ref{tab:occupancy}). The comparison is indirect and we do not lean on it: three of the four annotate sound events, not human actions, and two are synthesized scenes, dense because a generator fills a timeline, not because the world does.

The advantage does scale with emptiness, but only where calls are scarce. Over the \pooledVids\ recordings of Section~\ref{sec:repl}, occupancy ranges from $\pooledOccMin\%$ to $\pooledOccMax\%$, and at $\occCorrTightPct\%$ of calls the gain over uniform correlates negatively with it (Pearson $\occCorr$, $p=\occCorrP$; Spearman $\occCorrRank$), with leave-one-out confined to $[\occCorrLooLo,\occCorrLooHi]$ so no single recording carries it. The relation is a property of the scarce-call regime, not a general one: it weakens as calls grow more plentiful and has vanished by half the stream ($\occCorrLoose$), which is where uniform sampling is already close to exhaustive. We read it as consistent with the argument of this section, not as a law. Occupancy estimates headroom and no more: it says nothing about how long events run, whether they cluster, or whether they are audible, and Section~\ref{sec:repl} shows the first mattering on its own.

\subsection{Selecting which windows to call}
\label{sec:select}

\begin{figure}[t]
\centering
\includegraphics[width=\linewidth]{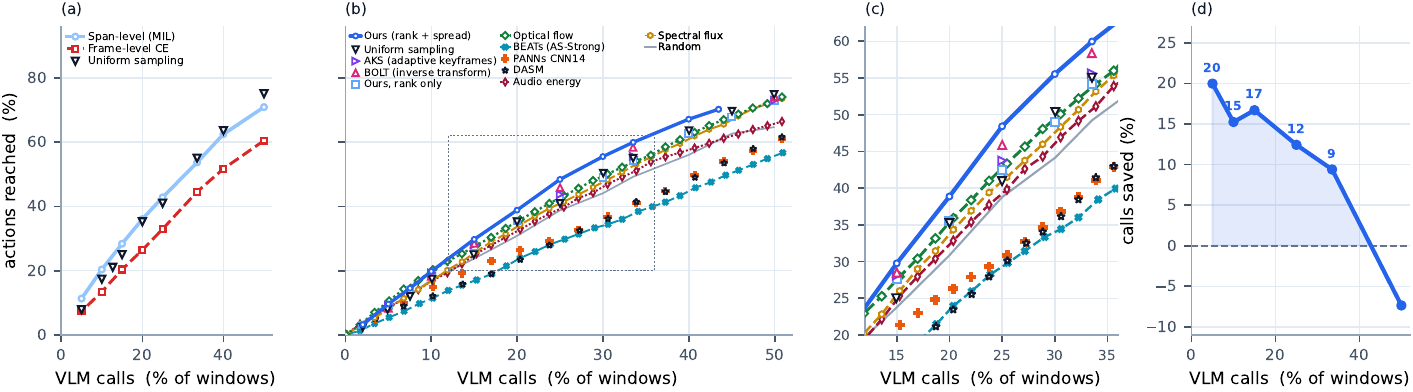}
\caption{EK-100, in VLM calls, below half the grid. (a) Training objective. (b) Every frontline and selection rule on one plane. (c) The boxed span at scale. (d) Calls saved against uniform sampling at equal coverage.}
\label{fig:callresults}
\end{figure}

Fig.~\ref{fig:callresults}(b) places every rule and frontline on one plane. Taking the top-scoring windows, the obvious use of a frontline, \emph{loses} to uniform sampling when calls are plentiful: a ranked list keeps returning to the same stretch, and repeat calls inside one action earn nothing.

The minimum separation of (\ref{eq:separation}) removes the redundancy while keeping the evidence, and recovers a consistent advantage below half the calls. Neither ingredient suffices alone: score without spread loses to uniform, and spread without score is uniform.

A hard separation has one structural defect: as the allowance approaches the number of admissible windows the rule forfeits calls, spending $1063$ of $1090$ at half the stream. A smooth coverage rule avoids this, which is why BOLT overtakes us there, as does a greedy facility-location variant of our own. Below half, where selections are already far apart, smoothing buys nothing and we report the simpler rule.

Points of coverage compare rules, but a deployment is sized the other way round: the allowance is fixed by a power envelope and the question is how much of it a required coverage consumes. Fig.~\ref{fig:callresults}(d) reads the same curves horizontally, as the calls uniform needs to match ours. The saving runs from $\saveMin\%$ near break-even to $\saveTightest\%$ at $\saveTightestCost\%$ of the stream and turns negative above half.

The same problem has been posed from the visual side. The selection stage of AKS~\cite{tang2025aks} consumes an arbitrary score sequence, so we run it on our acoustic scores. Two adaptations were needed, both favoring AKS: its split allocates $\text{budget}/2^{\text{depth}}$ windows per segment, which truncates to zero when calls are few, so we cap the depth at $\log_2$ of the allowance and top up any shortfall with the highest-scoring remaining windows. BOLT needs the same top-up, since a quantile can land twice in one window. Both spend theirs exactly at every level.

AKS behaves as its authors describe, improving on plain ranking and tracking uniform closely, which confirms the coverage term is what matters; we lead it by \aksGain\ points at \aksGainCost\% of the calls and converge at the tightest.

BOLT~\cite{liu2025bolt} takes the same treatment and is the stronger comparator: reading the score as a density and sampling at uniform quantiles concentrates calls where the score has mass without abandoning the rest of the recording. At the sharpness its authors recommend it leads AKS when calls are plentiful, and leads us at half the stream where a minimum separation starves ($\boltFifty\%$ against $70.2\%$); below half we lead by $\boltGainMin$ to $\boltGainMax$ points. The rules fail differently: theirs can place several calls inside one action, ours refuses the second call but runs out of legal windows.

\noindent\textbf{Replication on a second held-out set.}
\label{sec:repl}
Twelve recordings is a thin basis for margins of a few points, so we repeated the comparison on a second held-out set. The set was fixed before any of it was scored: from the EK-100 validation split we took the lowest-numbered recording of at least five minutes from each of \replVids\ participants the primary set never touches. That gives \replHours\ hours and \replActions\ actions. Eight of the twelve participants appear nowhere in training, four appear only through different recordings; the gain over uniform is $6.0$ points on the former and $5.6$ on the latter, so participant overlap is not what the margin is made of.

The set is not a copy of the first. Its actions run about twice as long ($\replActLen$\,s against $\primActLen$\,s), which suits uniform sampling: it reaches $53.5\%$ of actions at a quarter of the calls where it reached $40.9\%$ on the primary set. Our margin narrows accordingly but does not close, and the ordering of the rules is unchanged below half the stream.

Pooling both sets gives \pooledVids\ recordings from \pooledVids\ participants and \pooledActions\ actions, and because every rule runs on one recording at one allowance the comparison is paired, which removes the recording-to-recording spread that otherwise swamps a difference of a few points. Table~\ref{tab:paired} reports it at $\pairCallPct\%$ of calls. Our rule wins on \pairWinsUnif\ of \pooledVids\ recordings against uniform sampling, by $\pairGainUnif$ points on average ($p=\pairPUnif$, Wilcoxon signed-rank), and leads AKS and rank-only by a similar margin. BOLT remains the closest, at $+3.2$ points on 16 recordings of 24 ($p=0.019$). At half the stream the sign reverses against uniform sampling on 20 recordings of 24, which is the same starvation effect described above and the reason we do not claim that range.

\begin{table}[t]
\centering
\small
\caption{Paired comparison at $\pairCallPct\%$ of calls, on both datasets. W/L/T counts recordings or clips won, lost and tied.}
\label{tab:paired}
\input{tables/tab_paired.tex}
\end{table}

\label{sec:e2e}
Under the retained-duration account earlier work reports, the ordering is the expected one: our gate discards \gateRedOurs\% of the stream at \gateArOurs\% recall against \gateRedPanns\% for the AudioSet tagger. Much of that margin is an artifact of the unit, which is why we do not build on it.

\begin{table}[t]
\centering
\small
\caption{The same span-level head over five frozen extractors, at a $93\%$ development recall target.}
\label{tab:transfer}
\input{tables/tab_transfer.tex}
\end{table}

\subsection{The training objective, and what transfers}
\label{sec:gate}

Every comparison below uses an operating point selected on development recordings and applied unchanged to the evaluation recordings.

The open-vocabulary detector DASM~\cite{cai2025detectsoundopenvocabularysound} reaches \dasmRedNineThree\% reduction at $93\%$ recall, behind both the 2020 CNN baseline and our gate, despite a purpose-written kitchen vocabulary.

Three observations follow. The span-level objective can be decisive, roughly doubling the stream discarded at comparable recall relative to the threshold rule (\ruleRed\%) and to the AudioSet tagger (\gateRedPanns\%). The effect is not specific to our features, since the same head on PANNs posteriors improves substantially over PANNs thresholded directly, as expected if the objective, and not the representation, is doing the work. Further, the margin over the external baseline is large relative to seed variance ($\oursSeedRed\,\pm\,\oursSeedStd$\% over \nSeeds\ seeds against \gateRedPanns\%), whereas differences \emph{between} extractors are not (Table~\ref{tab:transfer}).

Most consequential for deployment, in-domain sound-event supervision buys nothing. A 42-class head on the domain's own taxonomy discards more stream at each dev target only by operating at lower recall: $62.6\%$ at $89.0\%$ against the cross-domain system's $64.3\%$ at $90.8\%$. Along the curve the two are equivalent within seed variance, so the cross-domain gate wins on the criterion that decides adoption.

A frame-level cross-entropy head trained on the same data with the same architecture underperforms even the hand-tuned threshold rule at high recall. Under matched architecture, features and training data, the dominant difference in these experiments is the training objective.

\label{sec:transfer}

If the gain came from our particular features, the reformulation would be of narrow interest. Trained with the identical recipe and protocol on four further extractors, the head helps in every case and separates none of them (Table~\ref{tab:transfer}): no pairwise comparison is significant ($p \ge 0.14$, Welch) and the per-seed ranges overlap almost entirely. The gain does not depend on a particular representation, and our measurements do not support choosing between these extractors on gating grounds.

Where they differ is stability: our head's recall varies by $\pm\seedSdOurs$ points across seeds, the tightest of the five, against $\pm\seedSdMin$ to $\pm\seedSdMax$ for the AudioSet-Strong extractors. A deployment commits to a recall target, so landing within a point of it is worth more than the same mean with more scatter. That, and being the only extractor able to place boundaries, is why we retain the domain-tuned head.

Individual runs are unstable. Repeating one configuration under identical data and hyper-parameters moved its reduction by more than $20$ points, so comparisons require the seed statistics we report throughout.

\label{sec:cross}
Applied without adaptation to Ego4D (\egoClips\ clips, \egoHours\,h~\cite{zhu2026egosound}) the gate is markedly weaker, retaining $80.0\%$ of spans at $44.5\%$ of the stream, so the representation transfers but the operating point does not.

\begin{figure}[t]
\centering
\includegraphics[width=\linewidth]{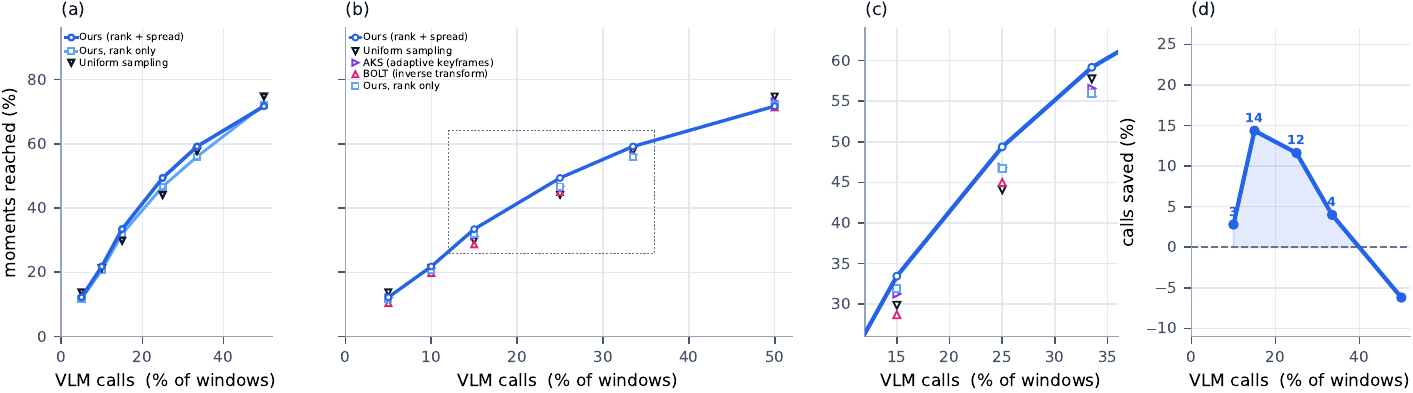}
\caption{Ego4D, in VLM calls, on the axes of Fig.~\ref{fig:callresults}.
(a) The selection rule isolated: score alone, spread alone, and both.
(b) Every rule on one plane. (c) The boxed span at scale. (d) Calls saved
against uniform sampling at equal coverage; the tightest rate is omitted, where
our coverage falls below anything uniform sampling reaches on this grid.}
\label{fig:ego4dpanels}
\end{figure}

\noindent\textbf{The selection rule on Ego4D.}
The comparison above is on one dataset, so we repeated it on Ego4D: the \egoSelClips\ of \egoClips\ clips carrying parseable spans and enough windows for a sweep, \egoSelHours\ hours and \egoSelMoments\ annotated moments, every rule run at matched cost and compared per clip. What is scored is not the same. Ego4D's annotations here are question-referenced acoustic moments rather than exhaustive action intervals, so the coverage values in Fig.~\ref{fig:ego4dpanels} are not comparable with EK-100's; the comparison \emph{between} rules is, since each is scored against the same targets on the same clip.

Ego4D is not an industrial dataset, but much of it is work: carpentry, fitting, retail counters, fabrication (Fig.~\ref{fig:datasets}). The ordering survives in the middle of the range and reverses at both ends. At a quarter of the calls our rule leads uniform sampling by \egoGainTwentyFive\ points on \egoWinsTwentyFive\ of \egoSelClips\ clips ($p=\egoPTwentyFive$); at half the stream it loses, as on EK-100 and for the same reason (Fig.~\ref{fig:ego4dpanels}). The tightest settings are different in kind. These clips are short, so $5\%$ of calls is \egoCallsFive\ call on the median clip, where evenly spaced sampling means the midpoint and a rule whose only mechanism is a minimum separation has nothing to act on: it degenerates to taking the single highest-scoring window, which loses to the midpoint by \egoLossFive\ points. We report it rather than trimming the range, but it measures the score, not the rule.

\label{sec:compose}\label{sec:placement}
\noindent\textbf{Composition and deployed naming accuracy.}
Recomputed in calls, triage and token compression~\cite{luo2025openmmego, shang2025prumerge} compose less usefully than duration suggests: compression alone matches the combination down to roughly $10\%$ of original cost, and only below that does the gate keep the system running, returning $11.5\%$ of actions at $2\%$ of cost against nothing. Naming accuracy differs by at most \indepMaxGap\ points between all actions and gate-retained actions across six token allowances. Every captioning number above is measured at the ground-truth action midpoint, which a deployment cannot locate; re-extracting the frame a deployment would select, object naming falls from $38.1\%$ to $35.0\%$ (exact McNemar $p=0.27$) and action naming is unchanged, so the VLM stage is the constant factor the end-to-end comparison assumes.

\subsection{Comparison with a visual frontline}
\label{sec:visual}

Video could instead be filtered with cheap visual motion cues. We test frame differencing and Farneb\"ack optical flow, and report the result as measured, including where it goes against us. Optical flow is the \emph{stronger} frontline at high recall, discarding \flowGapNineFive\ points more of the stream than we do at $95\%$ recall; the ordering reverses by $\crossRecall\%$, and at $87\%$ we lead by \oursGapEightSeven\ points. The common claim that ego-motion renders visual gating unusable in first-person video is not supported by these numbers.

Fig.~\ref{fig:callresults}(b) settles this in the unit that is billed, where the ordering above does not survive. Every off-the-shelf acoustic frontline sits well below uniform sampling, reaching $\frontAcMin$ to $\frontAcMax\%$ of actions at $\frontCallPct\%$ of the calls against $\frontUnif\%$, which is Section~\ref{sec:gate} seen from the cost side: the objective, not the representation, is what makes a frontline useful. Two baselines with no model at all sharpen this: short-term energy reaches $\frontEnergy\%$ and spectral flux $\frontFlux\%$, beating every pretrained detector and within a point of uniform sampling. A practitioner should try these first; we lead them by $7$ to $9$ points. Optical flow is the strongest alternative and tracks uniform closely, while we reach $\frontOurs\%$ against its $\frontFlow\%$, a lead of $\frontFlowGap$ points across the plotted range. The advantage flow holds at $93$ to $95\%$ recall is real but lies outside this range, at call rates where a system is no longer triaging.

Read as call savings it is closer still and in places reverses: at $5\%$ of the stream flow saves \flowSaveTight\% against our \oursSaveTight\%, and our margin appears only from mid range up (\oursSaveMid\% against \flowSaveMid\% at a quarter). Accuracy alone does not separate the two.

What does is cost. A visual frontline must decode video, the step we are trying to avoid: \visualRT$\times$ real time against \rtFactor$\times$ for audio, a factor of \visualSlowdown, or $117$ frontline GPU-hours per 1{,}000 against \gpuFront. Charged the same way, optical flow at its most favorable operating point costs $117 + 92 = 209$ GPU-hours, exceeding the \gpuFull\ of an unfiltered scan. A frontline whose own compute approaches the savings it provides is impractical whatever its accuracy.

\label{sec:seg}

One axis still favors a domain-tuned representation: at matched segment counts acoustic episodes reach $22.8\%$ $B\text{-}F_1@0.5$\,s against $17.4\%$ for uniform windows, and no AudioSet-Strong extractor places boundaries at all. 

\subsection{What does not work}
\label{sec:negative}

Fig.~\ref{fig:callresults}(a) makes the point in the billed unit: frame-level cross-entropy falls below uniform sampling at every operating point, so a frontline trained the conventional way is worse than not having one. Nor does the extractor rescue it: AudioSet-Strong heads used directly as the gate (BEATs, ATST-Frame, fPaSST) reduce the stream by $21.3\%$, $23.8\%$ and $16.3\%$ at $93\%$ recall, the first being the representation our gate reads, thresholded directly. Downsampling the discarded stream does not help: sampling a $2$\,s window every $\Delta$ seconds where the gate retained nothing looks worth $3$ to $5$ points when $\Delta$ is swept on the evaluation set, but under the dev-selected protocol the gain falls inside seed variance. Finally, hand-crop refinement and audio-conditioned prompting leave object naming unchanged (\capNounD\% against \capNounA\% for the full frame, $p=\capPAD$; and $p=\capPAB$): a hand detector localizes the hand, not the object.
\section{Discussion}
\label{sec:practice}

\textbf{When is triage worthwhile?}
Pre-decoding triage is most useful when a substantial fraction of the stream
can be avoided. Target-event occupancy provides a simple estimate of this
headroom. On the relatively dense public datasets in
Table~\ref{tab:occupancy}, uniform sampling already intersects many annotated
targets~\cite{damen2018epic,grauman2022ego4d}, leaving limited room for
content-aware selection. Occupancy does not capture event duration, temporal
clustering, or acoustic detectability, but it gives a useful first estimate of
the opportunity available beyond uniform sampling.

That opportunity matters only if the selector is cheap relative to the VLM
stage. Let $C_{\mathrm{full}}$ denote the cost of exhaustively applying the
VLM to a recording and $C_{\mathrm{gate}}$ the cost of screening the same
recording. If triage retains a fraction $\rho$ of the exhaustive VLM calls,
it reduces total compute when
\begin{equation}
C_{\mathrm{gate}} + \rho\,C_{\mathrm{full}} < C_{\mathrm{full}}
\quad\Longleftrightarrow\quad
\frac{C_{\mathrm{gate}}}{C_{\mathrm{full}}} < 1-\rho .
\label{eq:breakeven}
\end{equation}
In our implementation, acoustic screening costs about $3\%$ of an exhaustive
VLM scan, compared with about $75\%$ for optical flow. The acoustic gate
remains below this break-even threshold over the evaluated operating range,
whereas visual motion screening can consume much of the compute saved by
reducing downstream calls. This comparison is specific to selectors that
require decoded RGB frames or visual features before making the selection
decision.

The choice of cost unit can also change the comparison. Retained duration
underestimates the cost of fragmented selections when several short intervals still produce separate VLM calls. For our gate, this discrepancy
reaches $\costUndercount\times$. We therefore use VLM calls as the primary
cost measure and retain duration-based results only as a secondary view.

\textbf{What makes triage effective?}
Span-level supervision improves action coverage across all frozen audio
representations tested in Table~\ref{tab:transfer}, while differences between
extractors remain small relative to seed variation. The improvement is not
specific to the backbone used in the main experiments. Only the lightweight
$0.41$\,M-parameter temporal head is trained; the underlying audio
representation remains frozen.

The learned score still has to be converted into a useful allocation of calls.
Direct ranking repeatedly selects temporally correlated evidence, whereas
temporal spreading without a content score reduces to uniform sampling.
Minimum-separation selection works well where calls are scarce or moderate
because it combines coverage-aligned scores with temporal diversity. Where calls are plentiful, the hard separation constraint becomes restrictive, and smoother
strategies such as BOLT~\cite{liu2025bolt} can use the available calls more
fully. This behavior is consistent with prior visual selectors that explicitly
model temporal coverage or diversity~\cite{gong2014seqdpp,sun2025mdp3,
liu2025bolt}.

This distinction also explains why conventional detection metrics are not
sufficient for the deployment objective considered here. Frame- and
clip-level detectors~\cite{kong2020panns,schmid2024multi,10889422} are
optimized to respond whenever an event is present, whereas a gate under a call limit needs only enough evidence to place a useful call within an action. High
recall can therefore coexist with poor efficiency if it is achieved by
retaining most of the stream.

\textbf{Limitations and open directions.}
Most quantitative evidence comes from EK-100, while the VLM-based analyses use
a smaller subset because each retained stimulus requires a separate downstream
call. Whether a sparser operational regime exists, and whether a gate trained
in place would serve it, needs continuous recordings annotated at such a site;
that is the experiment we would run next.
Although the method does not require a domain-specific sound-event taxonomy,
training the temporal head still relies on annotated target-action spans.
Reducing this dependence is important for deployment in previously unseen
environments.

The method also relies on acoustic observability. Quiet manipulations, weak
sound signatures, or events masked by ambient noise may receive low gate
scores despite being visually salient. A sparse stream offers more headroom
only when its target events also provide usable acoustic evidence. Handling
such cases without giving up the pre-decoding advantage may require
lightweight fallback cues or adaptation mechanisms triggered only when audio
is uncertain.

Our primary coverage metric counts an action when a selected $w$-second window
overlaps its annotation. We also evaluate a stricter criterion that requires
the representative frame to fall inside the annotated interval, under which
the compared methods move closer to uniform sampling. These criteria capture
two levels of temporal alignment, but neither measures whether the downstream
VLM interprets the selected evidence correctly. Joint optimization of temporal
call placement and within-call visual sampling is therefore a natural next
step.

The reported audio--visual trade-off is specific to body-worn capture. A
head-mounted microphone records many manipulation sounds at close range,
whereas camera motion complicates inexpensive motion-based visual selection.
Static-camera video changes both conditions and may favor alternatives such as
background subtraction, so the reported margins should not be extrapolated
directly to fixed-camera settings.

Finally, the downstream analysis uses one VLM family, a fixed visual-input
protocol, and lexical noun/verb matching. The same VLM, prompt and token allowance are used across selection policies, but broader conclusions about
caption quality or factual correctness require additional VLMs and human
evaluation. More representative benchmarks with long continuous recordings and
naturally sparse target activity would also make it possible to test the
operating regime that motivates pre-decoding triage more directly.
\section{Conclusion}
\label{sec:conclusion}

We studied pre-decoding audio triage for VLM inference on long egocentric video. Rather than measuring efficiency by retained duration, we
evaluate cost directly in VLM calls and combine a span-level acoustic gate with a selection rule that respects the call limit. This aligns both learning and evaluation
with distinct-action coverage under a finite number of calls.

Span-level supervision improves action coverage by
$\objGainMin$--$\objGainMax$ percentage points over frame-level
cross-entropy at every operating point, with the same frozen audio
representation and temporal architecture. The improvement also holds across
the frozen representations we test without requiring domain-specific
sound-event labels. Direct score ranking, however, repeatedly selects
temporally correlated evidence; minimum-separation selection reduces this redundancy and improves coverage where calls are scarce.

The practical benefit depends on the recording regime and the cost of
screening. Dense public datasets leave limited headroom beyond uniform
sampling, while duration-based accounting can underestimate the number of VLM
calls induced by fragmented selections. The regime where triage should pay
most is the one public data does not contain. Pre-decoding audio triage is most
useful when target
activity is sufficiently sparse, acoustic evidence is informative, and
screening remains inexpensive relative to downstream VLM inference.

\appendix

\section{The paired comparison at every call rate}
\label{app:paired}

Table~\ref{tab:paired} reports one call rate because that is what the page
limit allowed. Tables~\ref{tab:apppairedepic} and~\ref{tab:apppairedego} give
the whole sweep, on the \pooledVids\ pooled EK-100 recordings and the
\egoSelClips\ Ego4D clips respectively. Each row is a Wilcoxon signed-rank test
over recordings or clips, paired because every rule runs on the same material
at the same allowance.

Two things are visible here that a single row cannot show. The advantage is not
a spike at one setting: against uniform sampling it holds from $33.5\%$ of calls
down to the tightest rate we run, and against rank-only and AKS it holds over
the same range with comparable margins. And the reversal at half the stream is
not a rounding artifact but the largest effect in either table, with uniform
sampling winning $20$ of $24$ recordings. That is the starvation described in
Section~\ref{sec:select}: as the allowance approaches the number of admissible
windows, a hard separation forfeits calls it is not permitted to place. We
report it at full strength rather than trimming the range.

BOLT is the closest comparator throughout, which is what we would expect of a
rule that spreads by construction rather than by constraint.

\begin{table}[t]
\centering
\small
\caption{EK-100, paired over \pooledVids\ recordings. Gain is our coverage minus
theirs, in points. W/L/T counts recordings won, lost and tied.}
\label{tab:apppairedepic}
\input{tables/tab_app_paired_epic.tex}
\end{table}

\begin{table}[t]
\centering
\small
\caption{Ego4D, paired over \egoSelClips\ clips, against question-referenced
acoustic moments rather than action intervals.}
\label{tab:apppairedego}
\input{tables/tab_app_paired_ego.tex}
\end{table}

\section{Every recording, at a quarter of the calls}
\label{app:perrec}

A mean of $\pairGainUnif$ points over \pooledVids\ recordings says nothing about
whether the margin is broad or carried by a few streams.
Table~\ref{tab:appperrec} lists all of them at $\pairCallPct\%$ of calls, with
the length and action statistics of each, so the gain can be read against the
spread it is drawn from.

The two recordings we lose are not the ones an outside reader would guess. The
margin does not track recording length, and it tracks mean action length only
weakly; what it tracks, as Section~\ref{sec:density} reports, is occupancy, and
only where calls are scarce. The replication set is visible in the table as the
block with longer actions and higher uniform-sampling coverage, which is why
our margin on it is the narrower of the two.

\begin{table}[t]
\centering
\small
\caption{All \pooledVids\ recordings at $\pairCallPct\%$ of calls. Act.\ is the
annotated action count, Mean the mean action length.}
\label{tab:appperrec}
\input{tables/tab_app_perrec.tex}
\end{table}

\section{Does a shorter window restore the advantage?}
\label{app:window}

Occupancy is a joint property of the recording and the call window: at $w=4$\,s
EK-100 is $\occupancy\%$ occupied, but a shorter window would contain an action
less often, sparsity would reappear, and a frontline that knows where the action
is ought to regain room. This is the first objection the occupancy argument
invites, so we tested it.

Table~\ref{tab:appwindow} sweeps $w$ from $0.5$ to $8$\,s. The rule scored here
is plain ranking by gate score, not the deployed minimum-separation rule, and
each entry is its coverage minus uniform sampling's at the same number of calls.
Occupancy behaves as expected, falling from $95.5\%$ at $8$\,s to $78.9\%$ at
half a second. The advantage does not follow it. Ranking is furthest behind
uniform sampling exactly where the stream is emptiest, by $17$ points at $0.5$\,s
and a quarter of the calls, and the deficit closes as windows lengthen.

The reason is the one Section~\ref{sec:problem} gives. Shortening the window
does not spread the evidence out; it subdivides the same acoustically active
stretches into more windows, all of which score highly and all of which a ranked
list will take before it looks elsewhere. Sparsity created this way is not
sparsity a score can exploit. It is an argument for the selection rule, not
against the window: what recovers the advantage is spreading the calls, at any
window length.

\begin{table}[t]
\centering
\small
\caption{Window length against ranking's coverage minus uniform sampling's, in
points, on the primary EK-100 set. The deployed setting is in bold.}
\label{tab:appwindow}
\input{tables/tab_app_window.tex}
\end{table}

\section{Operating points and training configuration}
\label{app:hparams}

Table~\ref{tab:apphparams} collects every constant the pipeline depends on.
Thresholds and the separation are chosen on development recordings and applied
unchanged to the evaluation recordings; nothing in the table is tuned on
anything it is later scored against. The three hysteresis thresholds correspond
to the three development recall targets we report throughout.

\begin{table}[t]
\centering
\small
\caption{Operating points and training configuration.}
\label{tab:apphparams}
\input{tables/tab_app_hparams.tex}
\end{table}

\bibliographystyle{IEEEtran}
\bibliography{references}

\end{document}

%% file: tables/macros.tex
\newcommand{\epicActions}{2208\xspace}
\newcommand{\egoClips}{260\xspace}

\newcommand{\egoHours}{7.45\xspace}
\newcommand{\rtFactor}{178\xspace}
\newcommand{\frameRate}{12.5\xspace}
\newcommand{\gateRate}{25\xspace}

\newcommand{\capNounA}{37.3\xspace}

\newcommand{\capNounD}{33.7\xspace}

\newcommand{\capPAD}{0.099\xspace}

\newcommand{\capPAB}{0.453\xspace}

\newcommand{\gateRedOurs}{53.9\xspace}
\newcommand{\gateArOurs}{94.3\xspace}

\newcommand{\gateRedPanns}{31.5\xspace}

\newcommand{\ruleRed}{19.8\xspace}

\newcommand{\gpuFull}{185\xspace}

\newcommand{\gpuFront}{6\xspace}
\newcommand{\gpuPerCall}{0.74\xspace}

\newcommand{\visualRT}{8.6\xspace}
\newcommand{\visualSlowdown}{21\xspace}
\newcommand{\crossRecall}{90\xspace}

\newcommand{\flowGapNineFive}{11.6\xspace}
\newcommand{\oursGapEightSeven}{3.8\xspace}
\newcommand{\selBest}{46.8\xspace}
\newcommand{\selBestUnif}{40.9\xspace}

\newcommand{\selBestCost}{25\xspace}
\newcommand{\selBestSd}{1.1\xspace}

\newcommand{\saveTightest}{20\xspace}
\newcommand{\saveTightestCost}{5\xspace}
\newcommand{\saveMin}{9\xspace}
\newcommand{\saveMax}{20\xspace}
\newcommand{\frontAcMin}{29\xspace}
\newcommand{\frontAcMax}{30\xspace}
\newcommand{\frontUnif}{41.0\xspace}
\newcommand{\frontOurs}{48.4\xspace}
\newcommand{\frontFlow}{42.7\xspace}
\newcommand{\frontFlowGap}{5.7\xspace}
\newcommand{\frontCallPct}{25\xspace}
\newcommand{\frontEnergy}{39.3\xspace}
\newcommand{\frontFlux}{40.9\xspace}
\newcommand{\flowSaveTight}{24\xspace}
\newcommand{\oursSaveTight}{17\xspace}
\newcommand{\flowSaveMid}{4\xspace}
\newcommand{\oursSaveMid}{14\xspace}
\newcommand{\objGainMin}{4.0\xspace}
\newcommand{\objGainMax}{10.8\xspace}
\newcommand{\occupancy}{85.1\xspace}
\newcommand{\actionsPerWindow}{1.41\xspace}

\newcommand{\occPubMin}{85.1\xspace}
\newcommand{\occPubMax}{99.3\xspace}
\newcommand{\aksGain}{3.8\xspace}
\newcommand{\aksGainCost}{34\xspace}
\newcommand{\boltGainMin}{0.6\xspace}
\newcommand{\boltGainMax}{1.7\xspace}
\newcommand{\boltFifty}{74.2\xspace}
\newcommand{\seedSdOurs}{0.7\xspace}
\newcommand{\seedSdMin}{1.0\xspace}
\newcommand{\seedSdMax}{2.5\xspace}
\newcommand{\oursSeedRed}{53.9\xspace}
\newcommand{\oursSeedStd}{7.3\xspace}
\newcommand{\nSeeds}{5\xspace}
\newcommand{\dasmRedNineThree}{24.6\xspace}
\newcommand{\epicEvalVids}{12\xspace}
\newcommand{\epicParticipants}{12\xspace}
\newcommand{\epicAudioHours}{2.42\xspace}
\newcommand{\gateTrainVids}{47\xspace}
\newcommand{\capStimuli}{200\xspace}
\newcommand{\indepMaxGap}{0.4\xspace}

\newcommand{\replVids}{12\xspace}
\newcommand{\replHours}{2.8\xspace}
\newcommand{\replActions}{1184\xspace}

\newcommand{\replActLen}{7.0\xspace}
\newcommand{\primActLen}{3.3\xspace}
\newcommand{\pooledVids}{24\xspace}
\newcommand{\pooledActions}{3392\xspace}
\newcommand{\pooledOccMin}{56.2\xspace}
\newcommand{\pooledOccMax}{99.2\xspace}
\newcommand{\pairCallPct}{25\xspace}
\newcommand{\pairGainUnif}{5.8\xspace}
\newcommand{\pairWinsUnif}{22\xspace}
\newcommand{\pairPUnif}{{<}10^{-4}\xspace}
\newcommand{\occCorr}{-0.59\xspace}
\newcommand{\occCorrP}{0.003\xspace}
\newcommand{\occCorrRank}{-0.59\xspace}
\newcommand{\occCorrLooLo}{-0.64\xspace}
\newcommand{\occCorrLooHi}{-0.45\xspace}
\newcommand{\occCorrTightPct}{5\xspace}
\newcommand{\occCorrLoose}{-0.00\xspace}
\newcommand{\egoSelClips}{247\xspace}
\newcommand{\egoSelMoments}{2234\xspace}
\newcommand{\egoSelHours}{7.4\xspace}
\newcommand{\egoGainTwentyFive}{5.2\xspace}
\newcommand{\egoWinsTwentyFive}{107\xspace}
\newcommand{\egoPTwentyFive}{{<}10^{-5}\xspace}
\newcommand{\egoLossFive}{1.8\xspace}
\newcommand{\egoCallsFive}{1\xspace}
\newcommand{\nCallsDeployed}{2{,}109\xspace}
\newcommand{\nDurEquiv}{1{,}201\xspace}
\newcommand{\nEpisodesDeployed}{2{,}809\xspace}
\newcommand{\meanEpisodeS}{1.7\xspace}
\newcommand{\costUndercount}{1.76\xspace}

%% file: tables/tab_methods.tex
\setlength{\tabcolsep}{3pt}
\begin{tabular}{@{}llc@{}}
\toprule
\textbf{System} & \textbf{What it scores} & \textbf{Ref.} \\
\midrule
\multicolumn{3}{@{}l}{\emph{Rules with no content score}}\\
Full scan            & every window                & --- \\
Uniform              & evenly spaced windows       & --- \\
Random spans         & matched length statistics   & --- \\
\addlinespace
\multicolumn{3}{@{}l}{\emph{Acoustic frontlines, thresholded}}\\
Short-term energy    & frame RMS                   & --- \\
Spectral flux        & onset strength              & --- \\
PANNs / CNN14        & AudioSet posteriors         & '20~\cite{kong2020panns} \\
BEATs, fPaSST, ATST-Frame & AudioSet-Strong frames & '22--'24~\cite{chen2022beats,koutini22passt,schmid2024multi} \\
DASM                 & open-vocabulary queries     & '25~\cite{cai2025detectsoundopenvocabularysound} \\
BEATs--SlowFast      & 42 in-domain sound classes  & '21~\cite{kazakos2021slowfast} \\
\addlinespace
\multicolumn{3}{@{}l}{\emph{Visual frontlines}}\\
Frame differencing   & pixel change                & --- \\
Optical flow, raw and ego-comp.& Farneb\"ack magnitude & --- \\
\addlinespace
\multicolumn{3}{@{}l}{\emph{Rules that spend a score}}\\
Rank-only            & top $K_{\max}$ by score     & --- \\
AKS                  & recursive split             & '25~\cite{tang2025aks} \\
BOLT                 & inverse transform sampling  & '25~\cite{liu2025bolt} \\
\textbf{Ours}        & span gate, min.\ separation & ours \\
\bottomrule
\end{tabular}

%% file: tables/tab_occupancy.tex
\setlength{\tabcolsep}{3pt}
\begin{tabular}{llccc}
\toprule
Dataset & Content & Recordings & Hours & Occupied (\%) \\
\midrule
EPIC-KITCHENS-100 & real daily activities & 24 & 5.5 & 85.1 \\
STARSS23 & real indoor scenes & 168 & 7.2 & 96.6 \\
TAU-NIGENS 2020 & synthesised scenes & 800 & 12.1 & 99.3 \\
DCASE 2022 synth & synthesised scenes & 1200 & 20.0 & 99.1 \\
\bottomrule
\end{tabular}

%% file: tables/tab_paired.tex
\setlength{\tabcolsep}{3pt}
\begin{tabular}{lcccc}
\toprule
Baseline & Mean $\Delta$ (pp) & Median $\Delta$ & W/L/T & $p$ \\
\midrule
\multicolumn{5}{l}{\textit{EK-100}, 24 recordings} \\
Uniform sampling & +5.8 & +5.2 & 22/2/0 & ${<}10^{-4}$ \\
Rank only (no spread) & +5.7 & +6.3 & 20/3/1 & ${<}10^{-4}$ \\
AKS~\cite{tang2025aks} & +5.6 & +5.6 & 20/1/3 & ${<}10^{-4}$ \\
BOLT~\cite{liu2025bolt} & +3.2 & +2.3 & 16/7/1 & $0.019$ \\
\midrule
\multicolumn{5}{l}{\textit{Ego4D}, 247 clips} \\
Uniform sampling & +5.2 & +0.0 & 107/59/81 & ${<}10^{-5}$ \\
Rank only (no spread) & +2.7 & +0.0 & 75/40/132 & ${<}10^{-3}$ \\
AKS~\cite{tang2025aks} & +2.8 & +0.0 & 78/38/131 & ${<}10^{-3}$ \\
BOLT~\cite{liu2025bolt} & +4.3 & +0.0 & 111/70/66 & ${<}10^{-3}$ \\
\bottomrule
\end{tabular}

%% file: tables/tab_transfer.tex
\setlength{\tabcolsep}{3pt}
\begin{tabular}{lccc}
\toprule
Feature extractor (our span-level head) & Recall (\%) & Reduction (\%) & Seed range \\
\midrule
PANNs CNN14~\cite{kong2020panns} & 94.1 $\pm$ 1.2 & 43.2 $\pm$ 4.7 & 36--51 \\
BEATs AudioSet-Strong~\cite{chen2022beats} & 95.6 $\pm$ 1.0 & 51.6 $\pm$ 1.8 & 50--55 \\
ATST-Frame~\cite{schmid2024multi} & 92.5 $\pm$ 2.5 & 56.3 $\pm$ 7.1 & 47--66 \\
fPaSST~\cite{koutini22passt} & 93.8 $\pm$ 1.5 & 52.6 $\pm$ 1.8 & 50--55 \\
BEATs-SlowFast (this work) & 93.8 $\pm$ 0.7 & 53.0 $\pm$ 2.6 & 50--57 \\
\bottomrule
\end{tabular}

%% file: tables/tab_app_paired_epic.tex
\begin{tabular}{@{}llrrrl@{}}
\toprule
\textbf{Calls} & \textbf{vs.} & \textbf{Gain} & \textbf{Median} & \textbf{W/L/T} & \textbf{$p$} \\
\midrule
50\% & Uniform & $-4.4$ & $-3.4$ & 2/20/2 & $2.9\times 10^{-4}$ \\
 & Rank-only & $-0.7$ & $-0.9$ & 10/13/1 & $0.693$ \\
 & AKS & $-0.5$ & $-0.7$ & 11/13/0 & $0.812$ \\
 & BOLT & $-2.9$ & $-2.5$ & 4/16/4 & $0.011$ \\
\addlinespace
33.5\% & Uniform & $+5.2$ & $+4.6$ & 19/3/2 & $4.3\times 10^{-4}$ \\
 & Rank-only & $+5.8$ & $+5.9$ & 21/3/0 & $<10^{-4}$ \\
 & AKS & $+5.7$ & $+4.6$ & 22/0/2 & $<10^{-4}$ \\
 & BOLT & $+1.9$ & $+0.0$ & 10/7/7 & $0.076$ \\
\addlinespace
25\% & Uniform & $+5.8$ & $+5.2$ & 22/2/0 & $<10^{-4}$ \\
 & Rank-only & $+5.7$ & $+6.3$ & 20/3/1 & $<10^{-4}$ \\
 & AKS & $+5.6$ & $+5.6$ & 20/1/3 & $<10^{-4}$ \\
 & BOLT & $+3.2$ & $+2.3$ & 16/7/1 & $0.019$ \\
\addlinespace
15\% & Uniform & $+4.7$ & $+4.7$ & 19/4/1 & $4.2\times 10^{-4}$ \\
 & Rank-only & $+2.4$ & $+2.3$ & 18/3/3 & $5.8\times 10^{-4}$ \\
 & AKS & $+1.7$ & $+1.7$ & 15/6/3 & $0.027$ \\
 & BOLT & $+0.2$ & $-0.1$ & 10/12/2 & $0.987$ \\
\addlinespace
10\% & Uniform & $+3.0$ & $+2.2$ & 17/5/2 & $0.006$ \\
 & Rank-only & $+1.8$ & $+1.8$ & 17/1/6 & $3.9\times 10^{-4}$ \\
 & AKS & $+0.9$ & $+0.6$ & 13/3/8 & $0.016$ \\
 & BOLT & $+3.5$ & $+1.4$ & 15/6/3 & $0.013$ \\
\addlinespace
5\% & Uniform & $+2.5$ & $+1.3$ & 16/6/2 & $0.004$ \\
 & Rank-only & $+0.4$ & $+0.0$ & 5/2/17 & $0.091$ \\
 & AKS & $+0.6$ & $+0.0$ & 11/7/6 & $0.267$ \\
 & BOLT & $+3.1$ & $+1.7$ & 16/5/3 & $0.004$ \\
\bottomrule
\end{tabular}

%% file: tables/tab_app_paired_ego.tex
\begin{tabular}{@{}llrrrl@{}}
\toprule
\textbf{Calls} & \textbf{vs.} & \textbf{Gain} & \textbf{Median} & \textbf{W/L/T} & \textbf{$p$} \\
\midrule
50\% & Uniform & $-3.2$ & $+0.0$ & 56/95/96 & $0.001$ \\
 & Rank-only & $-0.2$ & $+0.0$ & 83/81/83 & $0.917$ \\
 & AKS & $-1.4$ & $+0.0$ & 63/77/107 & $0.125$ \\
 & BOLT & $+0.5$ & $+0.0$ & 82/76/89 & $0.777$ \\
\addlinespace
33.5\% & Uniform & $+1.8$ & $+0.0$ & 94/82/71 & $0.211$ \\
 & Rank-only & $+3.5$ & $+0.0$ & 93/56/98 & $8.5\times 10^{-4}$ \\
 & AKS & $+2.9$ & $+0.0$ & 81/47/119 & $2.5\times 10^{-4}$ \\
 & BOLT & $+3.2$ & $+0.0$ & 99/66/82 & $0.002$ \\
\addlinespace
25\% & Uniform & $+5.2$ & $+0.0$ & 107/59/81 & $<10^{-4}$ \\
 & Rank-only & $+2.7$ & $+0.0$ & 75/40/132 & $9.3\times 10^{-4}$ \\
 & AKS & $+2.8$ & $+0.0$ & 78/38/131 & $2.7\times 10^{-4}$ \\
 & BOLT & $+4.3$ & $+0.0$ & 111/70/66 & $4.0\times 10^{-4}$ \\
\addlinespace
15\% & Uniform & $+3.3$ & $+0.0$ & 89/42/116 & $0.002$ \\
 & Rank-only & $+1.7$ & $+0.0$ & 41/15/191 & $0.002$ \\
 & AKS & $+2.4$ & $+0.0$ & 53/20/174 & $<10^{-4}$ \\
 & BOLT & $+4.5$ & $+0.0$ & 119/63/65 & $5.2\times 10^{-4}$ \\
\addlinespace
10\% & Uniform & $+0.3$ & $+0.0$ & 55/54/138 & $0.537$ \\
 & Rank-only & $+1.1$ & $+0.0$ & 21/7/219 & $0.003$ \\
 & AKS & $+0.6$ & $+0.0$ & 28/23/196 & $0.439$ \\
 & BOLT & $+1.6$ & $+0.0$ & 77/57/113 & $0.114$ \\
\addlinespace
5\% & Uniform & $-1.8$ & $+0.0$ & 34/55/158 & $0.032$ \\
 & Rank-only & $+0.5$ & $+0.0$ & 8/1/238 & $0.020$ \\
 & AKS & $+0.1$ & $+0.0$ & 3/1/243 & $0.144$ \\
 & BOLT & $+1.8$ & $+0.0$ & 53/34/160 & $0.004$ \\
\bottomrule
\end{tabular}

%% file: tables/tab_app_perrec.tex
\setlength{\tabcolsep}{3pt}
\begin{tabular}{@{}llrrrrrr@{}}
\toprule
\textbf{Recording} & \textbf{Set} & \textbf{Act.} & \textbf{Dur.\ (s)} & \textbf{Mean (s)} & \textbf{Unif.} & \textbf{Ours} & \textbf{$\Delta$} \\
\midrule
P01\_14 & primary & 354 & 1353 & 3.4 & 43.5 & 48.6 & $+5.1$ \\
P02\_12 & primary & 371 & 1315 & 3.1 & 42.0 & 46.1 & $+4.0$ \\
P03\_24 & primary & 136 & 822 & 3.9 & 39.0 & 61.0 & $+22.1$ \\
P04\_31 & primary & 113 & 825 & 5.7 & 50.4 & 54.9 & $+4.4$ \\
P05\_07 & primary & 111 & 802 & 3.6 & 49.5 & 55.9 & $+6.3$ \\
P08\_09 & primary & 147 & 621 & 3.4 & 42.2 & 44.2 & $+2.0$ \\
P11\_20 & primary & 195 & 560 & 2.4 & 35.4 & 42.1 & $+6.7$ \\
P18\_05 & primary & 135 & 594 & 3.6 & 47.4 & 54.8 & $+7.4$ \\
P20\_05 & primary & 126 & 467 & 3.1 & 42.9 & 50.0 & $+7.1$ \\
P22\_02 & primary & 217 & 509 & 2.3 & 34.1 & 40.6 & $+6.5$ \\
P28\_25 & primary & 133 & 397 & 2.4 & 36.1 & 40.6 & $+4.5$ \\
P30\_09 & primary & 170 & 434 & 2.4 & 34.1 & 43.5 & $+9.4$ \\
P10\_03 & replication & 235 & 1842 & 5.4 & 53.6 & 56.6 & $+3.0$ \\
P12\_03 & replication & 111 & 1466 & 8.9 & 59.5 & 60.4 & $+0.9$ \\
P13\_01 & replication & 49 & 554 & 8.6 & 63.3 & 65.3 & $+2.0$ \\
P14\_08 & replication & 33 & 398 & 4.8 & 45.5 & 60.6 & $+15.2$ \\
P15\_04 & replication & 37 & 471 & 7.9 & 62.2 & 67.6 & $+5.4$ \\
P16\_04 & replication & 57 & 930 & 11.3 & 70.2 & 77.2 & $+7.0$ \\
P17\_02 & replication & 27 & 546 & 9.3 & 63.0 & 74.1 & $+11.1$ \\
P21\_02 & replication & 60 & 491 & 6.6 & 58.3 & 56.7 & $-1.7$ \\
P23\_05 & replication & 104 & 1169 & 6.9 & 56.7 & 50.0 & $-6.7$ \\
P24\_09 & replication & 347 & 1969 & 4.4 & 47.0 & 51.9 & $+4.9$ \\
P27\_05 & replication & 64 & 615 & 5.3 & 43.8 & 51.6 & $+7.8$ \\
P32\_01 & replication & 60 & 495 & 4.5 & 51.7 & 56.7 & $+5.0$ \\
\midrule
\textbf{Mean} & & 141 & 818 & 5.1 & 48.8 & 54.6 & $+5.8$ \\
\bottomrule
\end{tabular}

%% file: tables/tab_app_window.tex
\setlength{\tabcolsep}{3pt}
\begin{tabular}{@{}rrrrrrrr@{}}
\toprule
\multicolumn{3}{@{}l}{} & \multicolumn{5}{c@{}}{\textbf{Calls (\% of grid)}} \\
\cmidrule(l){4-8}
\textbf{$w$ (s)} & \textbf{Windows} & \textbf{Occ.\ (\%)} & $50$ & $33.5$ & $25$ & $15$ & $10$ \\
\midrule
0.5 & 17402 & 78.9 & $-8.1$ & $-15.2$ & $-17.0$ & $-15.4$ & $-12.0$ \\
1 & 8703 & 83.0 & $-11.7$ & $-11.7$ & $-11.5$ & $-5.9$ & $-3.9$ \\
2 & 4354 & 88.0 & $-9.3$ & $-5.8$ & $-1.9$ & $-1.3$ & $-0.4$ \\
\textbf{4} & 2181 & 92.1 & $-1.8$ & $-0.8$ & $+1.5$ & $+2.6$ & $+1.3$ \\
8 & 1094 & 95.5 & $-0.5$ & $+2.4$ & $+2.4$ & $+2.8$ & $+1.6$ \\
\bottomrule
\end{tabular}

%% file: tables/tab_app_hparams.tex
\begin{tabular}{@{}llp{0.42\columnwidth}@{}}
\toprule
\textbf{Quantity} & \textbf{Value} & \textbf{Chosen by} \\
\midrule
\multicolumn{3}{@{}l}{\emph{Windowing and cost}} \\
Call window $w$ & $4$\,s & fixed, swept in Table~\ref{tab:appwindow} \\
Gate frame rate & $25$\,Hz & extractor \\
Chunk / hop & $10$\,s / $5$\,s & fixed \\
\addlinespace
\multicolumn{3}{@{}l}{\emph{Span-level objective}} \\
Pooling $\tau$ & $4.0$ & development recordings \\
Negative weight & $1.0$ & equal-weighted by construction \\
Optimizer & AdamW & --- \\
Learning rate & $3\times10^{-4}$ & development recordings \\
Weight decay & $1\times10^{-4}$ & development recordings \\
Epochs $\times$ steps & $15\times200$ & development recordings \\
Batch size & $16$ & fixed \\
Head size & $\approx0.2$\,M params & fixed \\
\addlinespace
\multicolumn{3}{@{}l}{\emph{Episode extraction}} \\
$\theta_{\mathrm{on}}$ at 99\% dev recall & $0.46$ & development recordings \\
$\theta_{\mathrm{on}}$ at 97\% dev recall & $0.62$ & development recordings \\
$\theta_{\mathrm{on}}$ at 95\% dev recall & $0.72$ & development recordings \\
$\theta_{\mathrm{on}}$ at 93\% dev recall & $0.76$ & development recordings \\
$\theta_{\mathrm{on}}$ at 90\% dev recall & $0.80$ & development recordings \\
$\theta_{\mathrm{on}}$ at 87\% dev recall & $0.84$ & development recordings \\
Median filter width & $1$ frame & development recordings \\
Minimum span & $0.15$\,s & shorter than any action \\
Maximum closed gap & $0.6$\,s & development recordings \\
\addlinespace
\multicolumn{3}{@{}l}{\emph{Selection}} \\
Separation $\delta$ & $2$ windows ($8$\,s) & development recordings \\
\bottomrule
\end{tabular}